\documentclass[11pt,a4paper]{article}
\usepackage[hyperref]{naaclhlt2019}
\usepackage{times}
\usepackage{latexsym}
\usepackage{graphicx}
\usepackage{adjustbox}
\usepackage{subfigure}
\usepackage{amsfonts}
\usepackage{url}
\usepackage{color}
\usepackage{tablefootnote}
\usepackage{multirow}
\usepackage{todonotes} 

\aclfinalcopy 

\DeclareGraphicsExtensions{.pdf,.png,.jpg}
\newcommand\MIN{SAN}

\title{Stochastic Answer Networks for Natural Language Inference}
\author{Xiaodong Liu$^\bold{\dagger}$, Kevin Duh$^\bold{\ddagger}$ and Jianfeng Gao$^\bold{\dagger}$ \\
  $^\bold{\dagger}$    
  Microsoft Research, Redmond, WA, USA \\
  $^\bold{\ddagger}$
  Johns Hopkins University, Baltimore, MD, USA \\
  {\tt $^\bold{\dagger}$\{xiaodl,jfgao\}@microsoft.com
   $^\bold{\ddagger}$kevinduh@cs.jhu.edu}
}
\date{}

\begin{document}
\maketitle
\begin{abstract}
We utilize a \textbf{s}tochastic \textbf{a}nswer \textbf{n}etwork (SAN) to explore multi-step inference strategies in Natural Language Inference. Rather than directly predicting the results given the inputs, the model maintains a state and iteratively refines its predictions. This can potentially model more complex inferences than the existing single-step inference methods.
Our experiments show that SAN achieves state-of-the-art results on four benchmarks: Stanford Natural Language Inference (SNLI), MultiGenre Natural Language Inference (MultiNLI), SciTail, and Quora Question Pairs datasets.
\end{abstract}
\section{Motivation}
\label{sec:mot}
The natural language inference task, also known as \textit{recognizing textual entailment} (RTE), is to infer the relation between a pair of sentences (e.g., \textit{premise} and \textit{hypothesis}).
This task is challenging, since it requires a model to fully understand the sentence meaning, (i.e., lexical and compositional semantics). For instance, the following example from MultiNLI dataset \cite{2017arXiv170405426W} illustrates the need for a form of multi-step synthesis of information between \textit{\textbf{premise}: ``If you need this book, it is probably too late unless you are about to take an SAT or GRE."}, and \textit{\textbf{hypothesis}: ``It's never too late, unless you're about to take a test."} To predict the correct relation between these two sentences, the model needs to first infer that ``\textit{SAT} or \textit{GRE}" is a ``\textit{test}", and then pick the correct relation, e.g., \textit{contradiction}. 

This kind of iterative process can be viewed as a form of multi-step inference. To best of our knowledge, all of works on NLI use a \textit{single step} inference. Inspired by the recent success of multi-step inference on Machine Reading Comprehension (MRC) \cite{hill2015goldilocks,dhingra2016gated,sordoni2016iterative,kumar15askme, liu18san, shen2017empirical,xu2018multi}, we explore the multi-step inference strategies on NLI.
Rather than directly predicting the results given the inputs, the model maintains a state and iteratively refines its predictions.
We show that our model outperforms single-step inference and further achieves the state-of-the-art on SNLI, MultiNLI, SciTail, and Quora Question Pairs datasets.
\section{Multi-step inference with SAN}
\label{sec:model}
\vspace{-0.1cm}
The natural language inference task as defined here involves a premise $P=\{p_0, p_1, ..., p_{m-1}\}$ of $m$ words and a hypothesis $H=\{h_0, h_1, ..., h_{n-1}\}$ of $n$ words,
and aims to find a logic relationship $R$ between $P$ and $H$, which is one of labels in a close set: \textit{entailment}, \textit{neutral} and \textit{contradiction}. 
The goal is to learn a model $f(P, H) \rightarrow R$. 

In a single-step inference architecture, the model directly predicts $R$ given $P$ and $H$ as input. 
In our multi-step inference architecture, we additionally incorporate a recurrent state $s_t$; the model processes multiple passes through $P$ and $H$, iteratively refining the state $s_t$, before finally generating the output at step $t=T$, where $T$ is an a priori chosen limit on the number of inference steps. 

\begin{figure*}[h!]
\centering
\adjustbox{trim={.081\width} {.01\height} {.05\width} {.01\height},clip}%
  {\includegraphics[scale=0.6]{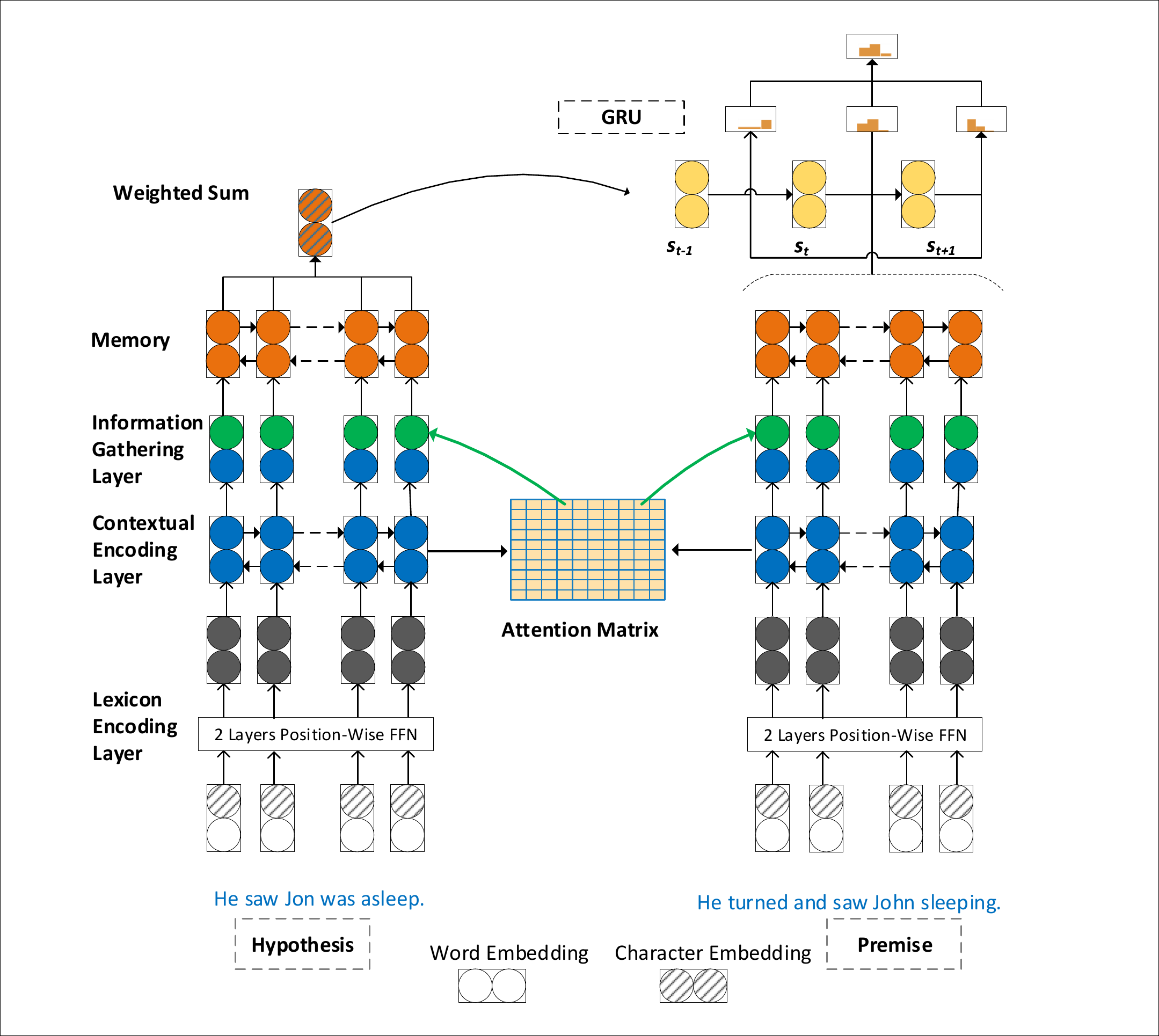}}
\caption{\label{fig:model} {Architecture of the Stochastic Answer Network ({\MIN}) for Natural Language Inference.}}
\end{figure*}

Figure \ref{fig:model} describes in detail the architecture of the \textbf{s}tochastic \textbf{a}nswer \textbf{n}etwork (SAN) used in this study; this model is \textit{adapted} from the MRC multi-step inference literature \cite{liu18san}. Compared to the original SAN for MRC, in the SAN for NLI we simplify the bottom layers and 
Self-attention layers since the length of the premise and hypothesis is short). We also modify the answer module from prediction a text span to an NLI classification label.  
Overall, it contains four different layers:
1) the lexicon encoding layer computes word representations; 
2) the contextual encoding layer modifies these representations in context; 
3) the memory generation layer gathers all information from the premise and hypothesis and forms a ``working memory" for the final answer module; 
4) the final answer module, a type of multi-step network, predicts the relation between the premise and hypothesis. 


\noindent\textbf{Lexicon Encoding Layer}. 
First we concatenate word embeddings and character embeddings to handle the out-of-vocabulary words\footnote{We omit POS Tagging and Name Entity Features for simplicity}.  
Following \cite{liu18san}, we use two separate two-layer position-wise feedforward network  \cite{vaswani2017attention} to obtain the final lexicon embedings, $E^p \in \mathbb{R}^{d \times m}$ and $E^h\in \mathbb{R}^{d \times n}$, for the tokens in $P$ and $H$, respectively. Here, $d$ is the hidden size.

\noindent\textbf{Contextual Encoding Layer}. 
Two stacked BiLSTM layers are used on the lexicon encoding layer to encode the context information for each word in both $P$ and $H$. Due to the bidirectional layer, it doubles the hidden size. We use a maxout layer \cite{goodfellow2013maxout} on the BiLSTM to shrink its output into its original hidden size. 
By a concatenation of the outputs of two BiLSTM layers, we obtain $C^p\in \mathbb{R}^{2d \times m}$ and $C^h\in \mathbb{R}^{2d \times n}$ as representation of $P$ and $H$, respectively.

\noindent\textbf{Memory Layer}. We construct our working memory via an attention mechanism. 
First, a dot-product attention is adopted like in \cite{vaswani2017attention} to measure the similarity between the tokens in $P$ and $H$.
Instead of using a scalar to normalize the scores as in \cite{vaswani2017attention}, we use a layer projection to transform the contextual information of both $C^p$ and $C^h$:
\begin{equation}
A=dropout(f_{attention}(\hat{C}^p, \hat{C}^h)) \in \mathbb{R}^{m \times n}\\
\label{eq:align}
\end{equation}
where $A$ is an attention matrix, and dropout is applied for smoothing. 
Note that $\hat{C^p}$ and $\hat{C^h}$ is transformed from $C^p$ and $C^h$ by one layer neural network $ReLU(W_3x)$, respectively. 
Next, we gather all the information on premise and hypothesis by: $U^p = \left[C^p; C^hA\right] \in \mathbb{R}^{4d \times m}$ and $U^h = \left[C^h ; C^pA^{\prime}\right] \in \mathbb{R}^{4d \times n}$.
The semicolon $;$ indicates vector/matrix concatenation; $A^{\prime}$ is the transpose of $A$.
Last, the working memory of the premise and hypothesis is generated by using a BiLSTM based on all the information gathered: $M^p=BiLSTM([U^p; C^p])$ and $M^h=BiLSTM([U^h; C^h])$.


\noindent\textbf{Answer module}.
Formally, our answer module will compute over $T$ memory steps and output the relation label. 
At the beginning, the initial state $s_0$ is the summary of the $M^h$: $s_0=\sum_j \alpha_j M^h_{j}$, where $\alpha_j = \frac{exp(\theta_2 \cdot M^h_j)}{\sum_{j'}exp(\theta_2 \cdot M^h_{j'})}$. At time step $t$ in the range of $\{1, 2, ..., T-1\}$, the state is defined by $s_t = GRU(s_{t-1}, x_t)$. 
Here, $x_t$ is computed from the previous state $s_{t-1}$ and memory $M^p$: $x_t=\sum_j\beta_j M^p_j$ and $\beta_j = softmax(s_{t-1}\theta_3 M^p)$. 
Following \cite{mou2015natural}, one layer classifier is used to determine the relation at each step $t \in \{0,1,\ldots,T-1\}$. 
\begin{equation}
P^r_t = softmax(\theta_4[s_t; x_t; |s_t - x_t|; s_t \cdot x_t]).
\label{eq:begin}
\end{equation}

At last, we utilize all of the $T$ outputs by averaging the scores:
\begin{equation}
P^r = avg([P_0^{r}, P_1^{r}, ..., P_{T-1}^{r}]).
\label{eq:avg}
\end{equation}\vspace{-0.1cm}
Each $P_t^{r}$ is a probability distribution over all the relations, $\{1,\ldots,|R|\}$. 
During training, we apply \emph{stochastic prediction dropout} before the above averaging operation. During decoding, we average all outputs to improve robustness. 

This stochastic prediction dropout is similar in motivation to the dropout introduced by \cite{srivastava2014dropout}. The difference is that theirs is dropout at the intermediate node-level, whereas ours is dropout at the final layer-level. Dropout at the node-level prevents correlation between features. Dropout at the final layer level, where randomness is introduced to the averaging of predictions, prevents our model from relying exclusively on a particular step to generate correct output. 

\section{Experiments}
\label{sec:exp}
\subsection{Dataset}
\label{sec:data}
Here, we evaluate our model in terms of accuracy on four benchmark datasets. \textbf{SNLI} \cite{bowman2015large} contains 570k human annotated sentence pairs, in which the premises are drawn from the captions of the Flickr30 corpus, and hypothesis are manually annotated. \textbf{MultiNLI} \cite{2017arXiv170405426W} contains 433k sentence pairs, which are collected similarly as SNLI. However, the premises are collected from a broad range of genre of American English. The test and development sets are further divided into in-domain (\textbf{matched}) and cross-domain (\textbf{mismatched}) sets.  
The Quora Question Pairs dataset \cite{wang2017bilateral} is proposed for paraphrase identification. It contains 400k question pairs, and each question pair is annotated with a binary value indicating whether the two questions are paraphrase of each other. 
\textbf{SciTail} dataset is created from a science question answering (SciQ) dataset. It contains 1,834 questions with 10,101 \textit{entailments} examples and 16,925 \textit{neutral} examples. Note that it only contains two types of labels, so is a binary task. 
\subsection{Implementation details}
\label{sec:imp}
The spaCy tool\footnote{https://spacy.io} is used to tokenize all the dataset and PyTorch is used to implement our models. We fix word embedding with 300-dimensional GloVe word vectors \cite{pennington2014glove}. For the character encoding, we use a concatenation of the multi-filter Convolutional Neural Nets with windows $1, 3, 5$ and the hidden size $50, 100, 150$.\footnote{We limit the maximum length of a word by 20 characters. The character embedding size is set to 20.} So lexicon embeddings are $d$=600-dimensions. The embedding for the out-of-vocabulary is zeroed. The hidden size of LSTM in the contextual encoding layer, memory generation layer is set to 128, thus the input size of output layer is 1024 (128 * 2 * 4) as Eq~\ref{eq:begin}. The projection size in the attention layer is set to 256. To speed up training, we use weight normalization \cite{salimans2016weight}. The dropout rate is 0.2, and the dropout mask is fixed through time steps \cite{gal2016theoretically} in LSTM. The mini-batch size is set to 32. Our optimizer is Adamax \cite{kingma2014adam} and its learning rate is initialized as 0.002 and decreased by 0.5 after each 10 epochs.

\begin{table}[t!]
\centering
\begin{tabular}{@{\hskip1pt}l || c | c  }
\hline
 & Single-step & SAN \\\hline
MultiNLI matched & 78.69 & \textbf{79.88} \\
MultiNLI mismatched & 78.83 & \textbf{79.91} \\ \hline 
SNLI & 88.32 & \textbf{88.73} \\ \hline
Quora & 89.67 & \textbf{90.70} \\ \hline
SciTail &85.46  & \textbf{89.35} \\ \hline
\end{tabular}
\caption{\label{tab:main} Comparison of single and multi-step inference strategies on MultiNLI, SNLI, Quora Question and SciTail \textbf{dev} sets.}
\end{table}


\subsection{Results}
\label{sec:result}
One main question which we would like to address is whether the multi-step inference help on NLI. 
We fixed the lower layer and only compare different architectures for the output layer:
\begin{enumerate}
\item \textit{Single-step}: Predict the relation using Eq~\ref{eq:begin} based on $s_0$ and $x_0$. Here, $x_0=\sum_j \alpha_j M^p_{j}$, where $\alpha_j = \frac{exp(w \cdot M^p_j)}{\sum_{j'}exp(w \cdot M^p_{j'})}$\footnote{For direct comparison, this has the same three lower layers as Fig. \ref{fig:model} and only changes the answer module.}.
\item \textit{SAN}: The multi-step inference model. We use 5-steps with the prediction dropout rate $0.2$ on the all experiments.
\end{enumerate} 


Table~\ref{tab:main} shows that our multi-step model consistently outperforms the single-step model on the dev set of all four datasets in terms of accuracy. For example, 
on SciTail dataset, SAN outperforms the single-step model by +3.89 (85.46 vs 89.35). 

We compare our results with the state-of-the-art in Table~\ref{tab:comp}. Our model achieves the best performance on SciTai and Quora Question tasks. For instance, SAN obtains 89.4 (vs 89.1) and 88.4 (88.3) on the Quora Question and SciTail test set, respectively and set the new state-of-the-art. On SNLI and MultiNLI dataset, ESIM+ELMo \cite{peters2018deep}, GPT \cite{radford2018improving} and BERT \cite{devlin2018bert} use a large amount of external knowledge or a large scale pretrained contextual embeddings. However, SAN is still competitive these models. On SciTail dataset, SAN even outperforms GPT.  
Due to the space limitation, we only list two top models.\footnote{See leaderboard for more information: https://www.kaggle.com/c/multinli-matched-open-evaluation, https//www.kaggle.com/c/multinli-mismatched-open-evaluation, https://nlp.stanford.edu/projects/snli, http://data.allenai.org/scitail/leaderboard/.}

We further utilize BERT as a feature extractor\footnote{We run BERT (the base model) to extract embeddings of both premise and hypothesis and then feed it to answer models for a fair comparison.} and use the SAN answer module on top of it. Comparing with \textit{Single-step} baseline, the proposed model obtains +2.8 improvement on the SciTail test set (94.0 vs 91.2) and +2.1 improvement on the SciTail dev set (96.1 vs 93.9). This shows the generalization of the proposed model which can be easily adapted on other models \footnote{Due to highly time consumption and space limitation, we omit the results using BERT on SNLI/MNLI/Quora Question dataset.}.

\begin{table}[t!]
\centering
\begin{tabular}{@{\hskip1pt}l @{\hskip1pt}||@{\hskip1pt} c |@{\hskip1pt} c  @{\hskip1pt}}
\hline
\multirow{2}{*}{Model}& \multicolumn{2}{c}{MultiNLI Test} \\ \cline{2-3}
&\multicolumn{1}{ @{\hskip1pt} c| @{\hskip1pt}}{Matched}& Mismatched   \\ \hline
DIIN\cite{2017arXiv170904348G} & 78.8 & 77.8\\ \hline
BERT\cite{devlin2018bert} & \textbf{86.7} & \textbf{85.9} \\ \hline
{\MIN} & 79.3 & 78.7 \\ \hline \hline
\multicolumn{3}{c}{ SNLI Dataset (Accuracy\%)} \\ \hline
ESIM+ELMo &\multicolumn{2}{c}{88.7} \\ \hline
GPT\cite{radford2018improving} & \multicolumn{2}{c}{\textbf{89.9}}\\ \hline
{\MIN} & \multicolumn{2}{c}{\textbf{88.7}} \\ \hline
\hline
\multicolumn{3}{c}{ Quora Question Dataset (Accuracy\%)} \\ \hline \hline
\cite{tomar2017neural} &  \multicolumn{2}{c}{88.4}\\ \hline
\cite{2017arXiv170904348G} &  \multicolumn{2}{c}{ 89.1} \\ \hline
{\MIN} &  \multicolumn{2}{c}{\textbf{89.4}} \\ \hline
\multicolumn{3}{c}{ SciTail Dataset (Accuracy\%)} \\ \hline \hline
\cite{scitail} &  \multicolumn{2}{c}{ 77.3} \\ \hline
GPT\cite{radford2018improving} & \multicolumn{2}{c}{88.3}\\ \hline
{\MIN} &  \multicolumn{2}{c}{\textbf{88.4}} \\ \hline



\end{tabular}
\caption{\label{tab:comp} Comparison with the state-of-the-art on MultiNLI, SNLI and Quora Question \textbf{test} sets.}
\end{table}

\begin{table}[ht!]
\centering
\begin{tabular}{@{\hskip1pt}l |@{\hskip1pt} l |@{\hskip1pt} c |@{\hskip1pt} c | c }
\hline
\multirow{2}{*}{Tag}& \multicolumn{2}{c|@{\hskip1pt}}{Matched}& \multicolumn{2}{c}{Mismatched} \\ \cline{2-5}
&Chen$^1$& {\MIN} & Chen$^1$& {\MIN} \\ \hline  \hline
Conditional &\textbf{100\%}  &65\%	&\textbf{100\%} &81\%\\ \hline 
Word overlap	&63 \% &\textbf{86\%} &76\% &\textbf{92\%}\\ \hline
Negation &75\%  &\textbf{80\%} &72\% &\textbf{79\%}\\ \hline
Antonym &50\% &\textbf{77\%} &58\% &\textbf{85\%}	\\ \hline 
Long Sentence	&67\% &\textbf{84\%} &67\% &\textbf{79\%}\\ \hline 
Tense Difference &\textbf{86\%}	 &75\% &\textbf{89\%} &83\%\\ \hline 
Active/Passive &88\%	&\textbf{100\%} &91\% &\textbf{100\%}\\ \hline 
Paraphrase&78\% &\textbf{92\%} &89\% &\textbf{92\%} \\\hline 
Quantity/Time&33\%	  &\textbf{53\%}& 46\% &\textbf{51\%} \\ \hline 
Coreference	 &\textbf{83\%}	 & 73\% &80\% &\textbf{84\%}\\ \hline 
Quantifier &74\%  &\textbf{81\%} & 77\% &\textbf{80\%}\\ \hline 
Modal &75\%	&\textbf{79\%} &76\% & \textbf{82\%} \\ \hline 
Belief &73\% &\textbf{77\%} &74\% & \textbf{78\%}\\ \hline \hline
\end{tabular}

\caption{\label{tab:error} Error analysis on MultiNLI. See  \cite{2017arXiv170708172N} for reference.}
\end{table}

\noindent \textbf{Analysis:}
How many steps it needs? We search the number of steps $t$ from 1 to 10. We observe that when $t$ increases, our model obtains a better improvement (e.g., 86.7 ($t=2$)); however when $t=5$ or $t=6$, it achieves best results (89.4) on SciTail dev set and then begins to downgrade the performance. Thus, we set $t=5$ in all our experiments.

We also looked internals of our answer module by dumping predictions of each step (the max step is set to 5). Here is an example\footnote{Its ID is id 144185n with \textbf{premise}
(And he said, What's going on?) and \textbf{hypothesis}	(I told him to mind his own business.)} from MutiNLI dev set. Our model produces total 5 labels (\textit{contradiction, neutral,	neutral, neutral, and neutral}) at each step and makes the final decision by voting \textbf{\textit{neutral}}. Surprising, we found that human annotators also gave different 5 labels: \textit{contradiction, neutral, neutral, neutral, neutral}. It shows robustness of our model which uses collective wise.

Finally, we analyze our model on the annotated subset\footnote{https://www.nyu.edu/projects/bowman/multinli/multinli\_1.0\\\_annotations.zip} of development set of MultiNLI. It contains 1,000 examples, each tagged by categories shown in Table~\ref{tab:error}. Our model outperforms the best system in RepEval 2017 \cite{2017arXiv171104289C} in most cases, except on ``Conditional" and ``Tense Difference" categories. 
We also find that SAN works extremely well on ``Active/Passive'' and ``Paraphrase'' categories. Comparing with Chen's model, the biggest improvement of SAN (50\% vs 77\% and 58\% vs 85\% on Matched and Mismatched settings respectively) is on the ``Antonym'' category.
In particular, on the most challenging ``Long Sentence" and ``Quantity/Time" categories, SAN's result is substantially better than previous systems. This demonstrates the robustness of multi-step inference.
\section{Conclusion}
\label{sec:con}
We explored the use of multi-step inference in natural language inference by proposing a stochastic answer network (SAN).
Rather than directly predicting the results (e.g. relation $R$ such as entailment or not) given the input premise $P$ and hypothesis $H$, SAN maintains a state $s_t$, which it iteratively refines over multiple passes on $P$ and $H$ in order to make a prediction. 
Our state-of-the-art results on four benchmarks (SNLI, MultiNLI, SciTail, Quora Question Pairs) show the effectiveness of this multi-step inference architecture. In future, we would like to incorporate the pertrained contextual embedding, e.g., ELMo \cite{peters2018deep} and GPT \cite{radford2018improving} into our model and multi-task learning \cite{liu2019mt-dnn}.


\bibliography{acl_snli}
\bibliographystyle{acl_natbib}
%

\end{document}